\documentclass[runningheads]{llncs}

\usepackage{multirow}
\usepackage{eccv}

\usepackage{eccvabbrv}

\usepackage{graphicx}
\usepackage{booktabs}

\usepackage[accsupp]{axessibility}  

\usepackage[pagebackref,breaklinks,colorlinks,citecolor=blue]{hyperref}
\usepackage{hyperref}

\usepackage{orcidlink}

\renewcommand{\paragraph}[1]{\vspace{3mm}\noindent\textbf{#1}}

\begin{document}

{\title{
PapMOT: Exploring Adversarial Patch Attack against Multiple Object Tracking}} 

\titlerunning{PapMOT: Exploring Adversarial Patch Attack against MOT}

\author{Jiahuan Long\inst{1,2} \and
Tingsong Jiang\inst{2}\and
Wen Yao\inst{2}\thanks{Corresponding authors.} \and Shuai Jia\inst{1} \and Weijia Zhang\inst{1}
\and Weien Zhou\inst{2} \and Chao Ma\inst{1}$^{\star}$ \and Xiaoqian Chen\inst{2}
}

\authorrunning{J.~Long et al.}

\institute{MoE Key Lab of Artificial Intelligence, AI Institute, Shanghai Jiao Tong University \and
Defense Innovation Institute, Chinese Academy of Military Science
\email{jiahuanlong@sjtu.edu.cn, tingsong@pku.edu.cn, wendy0782@126.com, jiashuai@sjtu.edu.cn, weijia.zhang@sjtu.edu.cn, weienzhou@outlook.com, chaoma@sjtu.edu.cn, chenxiaoqian@nudt.edu.cn }}
\maketitle

\begin{abstract}
Tracking multiple objects in a continuous video stream is crucial for many computer vision tasks. It involves detecting and associating objects with their respective identities across successive frames. Despite significant progress made in multiple object tracking (MOT), recent studies have revealed the vulnerability of existing MOT methods to adversarial attacks. Nevertheless, 
all of these attacks belong to digital attacks that inject pixel-level noise into input images, and are therefore ineffective in physical scenarios.
To fill this gap, we propose PapMOT, which can generate \textbf{p}hysical \textbf{a}dversarial \textbf{p}atches against MOT for both digital and physical scenarios. Besides attacking the detection mechanism, PapMOT also optimizes a printable patch that can be detected as new targets to mislead the identity association process. Moreover, we introduce a patch enhancement strategy to further degrade the temporal consistency of tracking results across video frames, resulting in more aggressive attacks. We further develop new evaluation metrics to assess the robustness of MOT against such attacks. Extensive evaluations on multiple datasets demonstrate that our PapMOT can successfully attack various architectures of MOT trackers in digital scenarios. We also validate the effectiveness of PapMOT for physical attacks by deploying printed adversarial patches in the real world. 

\end{abstract}

\section{Introduction}

\begin{figure}[t]
\centering
\includegraphics[scale=0.255]{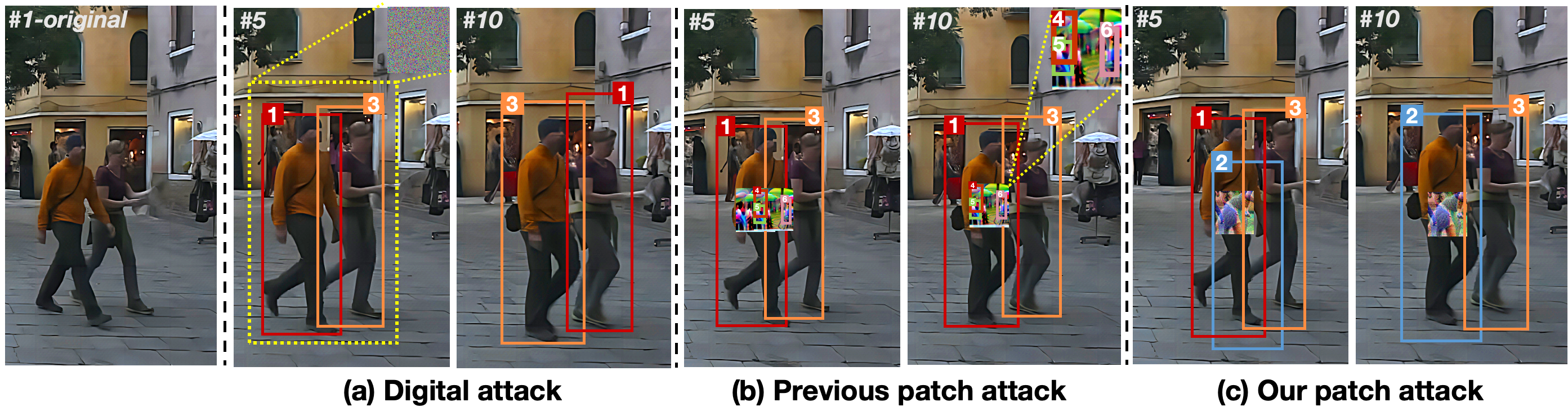}
\caption{
A comparison of various adversarial attack methods against MOT. (a) shows a digital attack~\cite{lin2021trasw} against MOT, where pixel-level noise is added to each frame and the identities of two people are switched within ten frames. (b) shows a patch attack~\cite{thys2019fooling} that fools only the detectors of MOT, generating new fake identities. (c) demonstrates that our patch attack simultaneously fools both the detectors and trackers of MOT, which creates new identities and alters the identities of persons as well.
}
\label{fig:different method comparison}
\end{figure}

Multiple object tracking (MOT) becomes increasingly widespread with numerous applications including autonomous driving \cite{Chiu_Li_Ambrus_Bohg_2021} and intelligent surveillance systems \cite{luo2021multiple}. Meanwhile, 
recent studies~\cite{lin2021trasw, hijack} have shown that deep multiple object tracking algorithms are susceptible to attacks by adversarial examples. These examples are designed to fool MOT models by adding imperceptible perturbations to video sequences. If an attack is successful, the identities of detected objects are no longer associated with their original identities. Existing adversarial attacks~\cite{lin2021trasw, hijack}  against MOT have mainly focused on digital attacks, manipulating the pixels of the entire image to fool the trackers. Although these digital attacks can craft imperceptible perturbations, they are generally ineffective in physical scenarios. For instance, Figure~\ref{fig:different method comparison}(a) illustrates an adversarial instance~\cite{lin2021trasw} against MOT that cannot be physically implemented. Compared to digital attacks, launching attacks against MOT in the physical world is more challenging. It requires strong perturbations that can withstand diverse environmental conditions, such as illumination and geometric alterations. Also, physical attacks are inherently more challenging to detect and defend against, making them more significant threats to detectors and trackers. Therefore, developing a physical attack against MOT is of great importance to the disclosure and understanding of MOT systems' vulnerabilities in real-world applications.

Physical adversarial patches are the most common form of physical attacks in the existing literature. They are widely utilized to gauge the reliability and robustness of computer vision systems, such as for face recognition~\cite{yang2020design, hu2021naturalistic} and object detection~\cite{advtexture, du2022physical}. However, applying these methods to MOT systems presents new technical challenges. A major requirement of successful MOT systems is to maintain object identity consistency across frames. This requires the detector to identify targets in each frame and the tracker to associate these targets frame-by-frame. Simply applying adversarial patches generated by detector-based attacks to the MOT system does not sufficiently disrupt data associations across frames. As shown in Figure~\ref{fig:different method comparison}(b), applying the patch attack~\cite{thys2019fooling} can create new fake identities but fails to alter the identity of the real attack object with ``ID = 1''. Moreover, the existing evaluation metrics of MOT attacks are not enough to reflect the performance of patch attacks. Existing metrics neglect the importance of the detection phase, which is a critical component of MOT algorithms. Also, these metrics only focus on the effects on one or a few targets, ignoring how attacks on these targets could affect the whole system.

To address these problems, we propose PapMOT, a physical adversarial patch scheme against MOT systems. PapMOT generates printable adversarial patches that create new identities in the tracking phase, thereby disrupting the existing identity associations. As depicted in Figure~\ref{fig:different method comparison}(c), PapMOT deceives high-performance detectors into outputting false detection boxes (\ie, blue detection boxes), overlapping with true detection boxes. This overlap presents a significant threat to the tracking system, causing it to mistakenly alter the identities of targets in a few frames. 
To this end, we introduce three adversarial losses that respectively consider the size of bounding boxes, scores of detection boxes, and the smoothness of color transitions. Besides, we propose a patch enhancement strategy that crops targets over time to obtain an extra dataset. The joint training of the original training dataset and the clip dataset will increase the attack success rate on specific targets.
To better evaluate the effectiveness of the attack, we construct maximum-attack-threshold evaluation metrics that integrate detection and tracking to comprehensively evaluate the performance of MOT attacks. We summarize our contributions as follows.

\begin{itemize}
\itemsep 0.5mm




\item To the best of our knowledge, this is the first work to explore physical adversarial attack against MOT. The proposed PapMOT attack is able to produce printable patches that successfully disrupt the data association process in tracking.

\item Upon delving into the association algorithm in MOT, we propose a novel bounding box restriction loss that generates a wrong detection box with a high IoU value, leading to incorrect identity assignment.

\item We propose a new set of metrics to evaluate attacks against MOT that is more comprehensive than previous MOT attacks.

\item We evaluate PapMOT in both digital and physical scenarios and demonstrate its favorable attack performance compared to state-of-the-art approaches across multiple datasets.
\end{itemize}


\section{Related Work}

\subsection{Multiple Object Tracking}
Existing MOT methods can be classified into two main categories: Tracking-by-Detection~\cite{luo2021multiple, bytetrack, deepsort, cao2022observation} and Joint Detection and Tracking~\cite{fairmot, zhao2022tracking}. In the former paradigm, object detectors are used to detect targets in each frame, and data association algorithms are then employed to link the detection results across video frames, including ByteTrack~\cite{bytetrack}, DeepSORT~\cite{deepsort}, and OCSort~\cite{cao2022observation}. ByteTrack~\cite{bytetrack} categorizes all detections into high and low scores and associates them with predicted boxes using the Kalman filter. This improves tracking performance in cases of occlusion and motion blur problems. In the latter paradigm, object detection and data association tasks are integrated into a single framework. FairMOT~\cite{fairmot} introduces an \textit{anchor-free} detection head, which allows its network to predict the position and size of the targets and generate a unique ID feature to achieve data association. In this work, we use ByteTrack~\cite{bytetrack} and FairMOT~\cite{fairmot} as representative trackers to validate the effectiveness and generalization of the proposed adversarial patch attack.

\begin{table}[t]
\caption{A comparison of representative adversarial attack methods. }
\small
\begin{center}
\setlength{\tabcolsep}{3pt}
\begin{tabular}{@{}ccccccc@{}}
\toprule
Methods  
 & \begin{tabular}[c]{@{}l@{}} NatPatch\\\quad~\cite{hu2021naturalistic}\end{tabular}   
 &\begin{tabular}[c]{@{}l@{}} ObjPatch\\\quad~\cite{thys2019fooling}\end{tabular}   
 &\begin{tabular}[c]{@{}l@{}} AdvTshirt\\\quad~\cite{xu2020adversarial}\end{tabular} 
 &\begin{tabular}[c]{@{}l@{}} Trasw\\~\cite{lin2021trasw}\end{tabular} 
&\begin{tabular}[c]{@{}l@{}} Hijack\\~\cite{hijack}\end{tabular} 
 &\begin{tabular}[c]{@{}l@{}} 
PapMOT\\\quad(ours)\end{tabular}  \\
\midrule
\ \  Advanced detectors 	&	 &	&      &     & 	 &$\surd$\\ 
\ \   Detectors \& trackers &	 &	&      &$\surd$     &$\surd$ 	 &$\surd$\\ 
\ \  Diverse datasets &$\surd$	 &	&$\surd$      &$\surd$    & 	 &$\surd$ \\ 
\ \  Physical attack  &$\surd$	 &$\surd$	&$\surd$      &    & 	 &$\surd$ \\ 
\ \  New evaluation metrics &	 &	&      &    & 	 &$\surd$ \\ \bottomrule
\end{tabular}
\end{center}
\label{Tab: different patch attacks} 
\end{table} 

\subsection{Adversarial Patch Attacks}
Previous digital attacks~\cite{xu2020adversarial1, jia2020robust, jia2021iou, jia2022exploring} are difficult to implement in physical scenarios as the perturbations are imperceptible for visual systems. To achieve physical attack, 
Thys et al. \cite{thys2019fooling} generate physically feasible adversarial patches for deceiving person detectors. Expanding upon this approach, Hu et al. \cite{advtexture} develop AdvTexture by printing the patch on a large cloth and making different T-shirts, skirts, and dresses to realize physical adversarial attacks on pedestrian detection. These two schemes enhance the robustness of patches in the real world by involving patch transformation \cite{athalye2018synthesizing}. However, they mainly conduct attacks on lower-precision detectors such as YOLOv2 \cite{Redmon_Farhadi_2017} and YOLOv3 \cite{Redmon_Farhadi_2018}.
For attacks on consecutive frames, 
%
Jia et al. \cite{hijack} propose a tracker hijacking attack scheme, which allows adversarial samples to successfully add or remove objects within a single frame, potentially posing a safety hazard for autonomous vehicles. 
Lin et al. \cite{lin2021trasw} present a Tracklet-Switch method, which introduces noise into certain frames, causing an ID-switch when pedestrians cross paths with each other. Yet, to our best knowledge, there has been no successful physical attack against MOT systems. 

To fill this research gap, we propose PapMOT, the first physically realizable attack method against MOT, alongside a set of newly proposed evaluation metrics. Different from all existing works, PapMOT is capable of not only attacking more advanced detectors (\ie, YOLOX), but also crippling both the detector and the tracker of an MOT system simultaneously and across a diversity of datasets and scenarios (\ie, MOT15, MOT17, MOT20, and BDD100K). Our contributions are expected to bring new insights into existing research on the digital and physical robustness of MOT systems. A comparison of the features and contributions of our method against prior works is provided in Table~\ref{Tab: different patch attacks}. 
%

\section{Methodology}

\begin{figure*}[t]
	\begin{minipage}{\linewidth}
		\centering
		\includegraphics[width=1.0\linewidth]{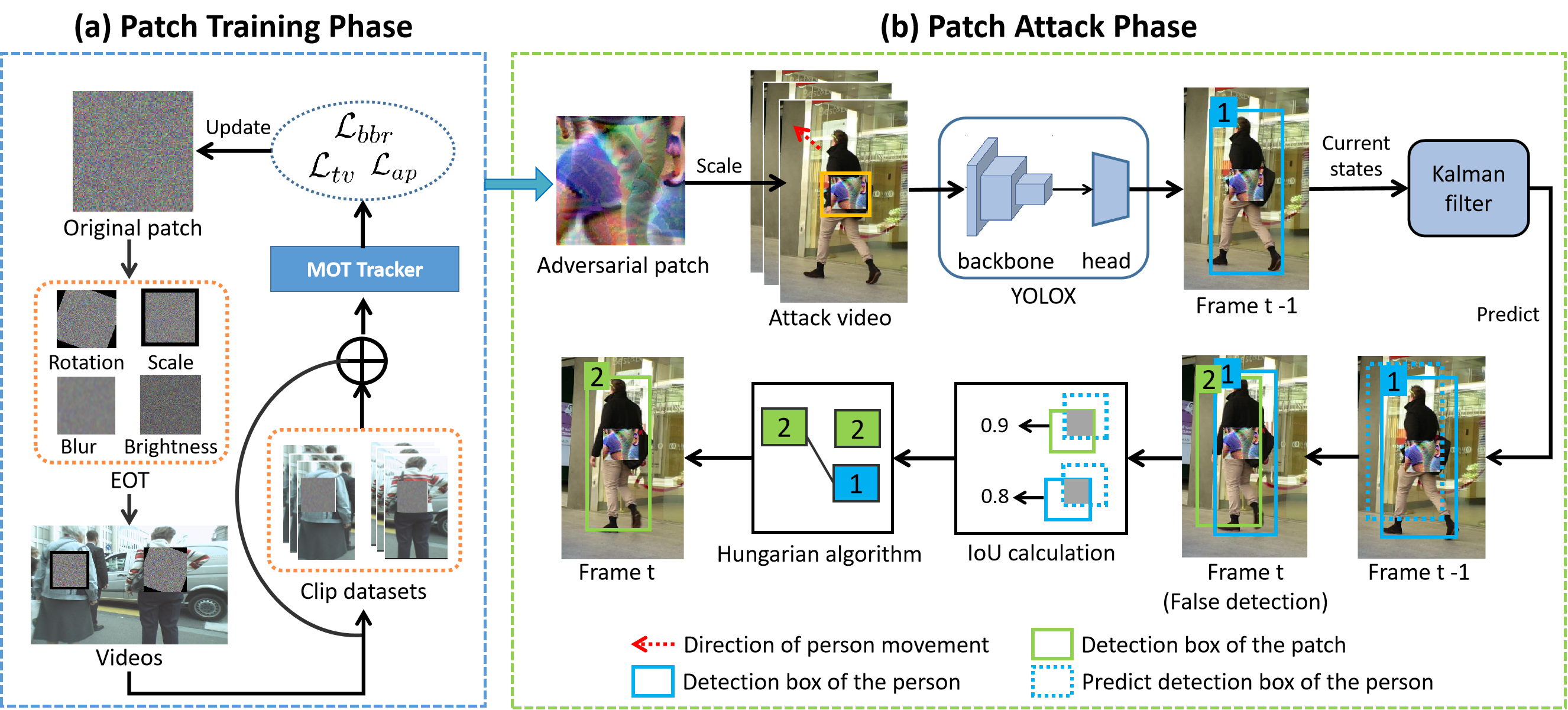}
	\end{minipage}
	\caption{The overall framework of PapMOT. (a) In the patch training phase, the objective is to optimize an adversarial patch to achieve effective attacks in both digital and physical domains. During this phase, Expectation over Transformation (EOT) is employed as data augmentation for patches. (b) Subsequently, the well-trained adversarial patch is deployed  in the patch attack phase to disrupt object detection and data association in MOT.}
\label{fig:optimazation framework} 
\end{figure*}

\subsection{Overall Framework of PapMOT}

An MOT system typically consists of a detection module and an association module. The association module generally employs IoU matching to maintain consistent IDs of objects across consecutive frames, which is at the core of high-performance MOT. Provided high IoU values of a detection box across adjacent frames, this box is likely to be assigned the same ID across these frames. As such, the core objective of our PapMOT is to train printable patches that can induce false detection boxes with a high IoU value with a correctly detected pedestrian box, leading to incorrect identity assignment of this very detection. The entire pipeline of the proposed PapMOT attack is depicted in
Figure~\ref{fig:optimazation framework}.


\paragraph{Patch Training Phase.} 
The patch training phase is illustrated in Figure~\ref{fig:optimazation framework}(a). First,  we randomly initialize a patch and employ a data enhancement strategy (i.e., EOT~\cite{athalye2018synthesizing}) to enhance its robustness.
EOT incorporates potential patch transformations such as rotation, brightness, blurring, and scaling that commonly exist in the physical world into patch training. Such transformations make our patch attacks more feasible in real-world situations. 
%
Next, we embed the EOT-enhanced patch into clean videos to obtain an adversarial dataset, and crop out pedestrian images based on their bounding boxes to produce an adversarial clip dataset.
%
Finally, we combine these two adversarial datasets to optimize the physical adversarial patch using three losses. We use total variation loss ($\mathcal{L}_{tv}$) to ensure patch smoothness and printability, bounding box restriction loss ($\mathcal{L}_{bbr}$) to minimize the discrepancy between the detection boxes of the patch and the object, and average precision loss ($\mathcal{L}_{ap}$) to decrease the detection confidence of targets with the patch. 
%

\paragraph{Patch Attack Phase.} 
Our patch can deceive detectors into outputting false detection boxes, thereby disrupting the identity association in tracking. This specific attack process is diagrammed in Figure~\ref{fig:optimazation framework}(b). The adversarial patch, already well-trained during the patch training phase, is embedded into a video to create an attack video. Next, the attack video is processed frame-by-frame through YOLOX, which outputs detection boxes for each frame. 
To clarify the disruption of data association, we consider frames $t-1$ and $t$ as illustrative examples. 
Assuming that there is a pedestrian with ``id=1'' in frame $t-1$, the Kalman filter~\cite{kalman_filter} outputs a predicted detection box (\ie, blue dashed-line box) by inputting the current motion states of the pedestrian. In the next frame $t$, two types of detection boxes appear, with one being a false detection box of the patch (\ie, green solid-line box) and the other being the detection box of the pedestrian (\ie, blue solid-line box). These two detection boxes do Intersection over Union (IoU) calculation with the predicted detection box from frame $t-1$. Following this, the Hungarian algorithm~\cite{hungarian_algorithm} uses these IoU values to assign identities to the detection boxes, leading the disruption of data association. (\ie, The false detection box of the patch in frame $t$ is matched with the identity of pedestrians from frame $t-1$).

\subsection{Adversarial Loss for Patch Attack}
\label{loss}
For our loss formulation, we adopt the widely used average precision loss and total variation loss~\cite{Sharif_Bhagavatula_Bauer_Reiter_2016, hu2021naturalistic, FCA} to reduce the confidence score of detection while encouraging visual smoothness of the printable patches. However, only using these two losses cannot disrupt the identity association in tracking, as illustrated Figure~\ref{fig:different method comparison}(b).
This motivates us to propose a novel bounding box restriction loss, which generates a wrong detection box with a high IoU value, leading to incorrect identity assignment and successfully disrupted data association.



\paragraph{Bounding Box Restriction Loss ($\mathcal{L}_{bbr}$).} To effectively disrupt identity association across frames, it is essential to ensure that the sizes of the target detection box and the patch detection box are similar. To this end, a bounding box restriction loss $\mathcal{L}_{bbr}$ is formulated to restrict the height and width of the detection box, while minimizing the discrepancy between the detection boxes of the patch and the object. $\mathcal{L}_{bbr}$ is defined as:
\begin{equation}
\mathcal{L}_{bbr} = \frac{1}{N}\sum_{i=1}^{N} (w_i/I_{width} + h_i/I_{height})  + (1-\frac{1}{N}\sum_{i=1}^{N} IoU(P_i, T_i)),
\end{equation}
where $N$ is the total number of attacked objects; $w_i$ and $h_i$ are the width and height of predicted bounding boxes of the $i$-th target; and $I_{width}$ and $I_{height}$ are the width and height of video frames that are used for normalization; detection boxes of the patch and the target are denoted as $P_i$ and $T_i$, respectively. 
It is worth noting that the first term seems to make $T_{i}$ have zero width and height, 
but we empirically find that it do not completely vanish and even create more new detection boxes due to the balance of IoU losses.

\paragraph{Total Variation Loss ($\mathcal{L}_{tv}$).} The aim of $\mathcal{L}_{tv}$ is to ensure the generation of printable patches with smooth color transitions, which helps to reduce the presence of visual noise. $\mathcal{L}_{tv}$ is formulated as:
\begin{equation}
\mathcal{L}_{t v}=\sum_{i, j} \sqrt{\left(\left(I_{i, j}-I_{i+1, j}\right)^{2}+\left(I_{i, j}-I_{i, j+1}\right)^{2}\right.},
\end{equation}
where $I_{i, j}$ denotes the pixel value at position ${(i, j)}$ in the patch. 

\paragraph{Average Precision Loss ($\mathcal{L}_{ap}$).} 
It is essential to lower the confidence of the original detection box in order to have it swapped to the one induced by the patch. $\mathcal{L}_{ap}$ does so exactly by reducing the average detection score of targets that have been attached to an adversarial patch. Given a set of $N$ specific detection boxes, where the detection score for the $i$-th box is $s_i$, $\mathcal{L}_{ap}$ is calculated as the sum of detection scores divided by the total number of detection boxes as follows:
\begin{equation}
\mathcal{L}_{ap} = \frac{1}{N} \sum_{i=1}^{N} s_i.
\end{equation}
Finally, the overall adversarial loss $L_{adv}$ is a sum of the above three loss terms weighted by hyperparameters $\beta$, $\gamma$, and $\delta$:
\begin{equation}
    \mathcal{L}_{adv}=\beta \mathcal{L}_{bbr}+\gamma \mathcal{L}_{tv}+\delta \mathcal{L}_{ap}
\label{total loss},
\end{equation}

\subsection{Evaluation Metrics for MOT Attack}
\label{valuation Metrics for MOT Attacks}

Current attack methods against MOT~\cite{hijack, lin2021trasw} have limitations in their evaluation since they utilize identity-based evaluation metrics. These metrics have two main drawbacks. Firstly, they are primarily designed for gauging tracking performance, thereby largely overlooking the quality of detection -- the other core ingredient in successful MOT. 
Secondly, they tend to focus only on a few specific targets, disregarding a comprehensive evaluation of the entire MOT system. 

\begin{figure}[t]
\centering
\includegraphics[scale=0.22]{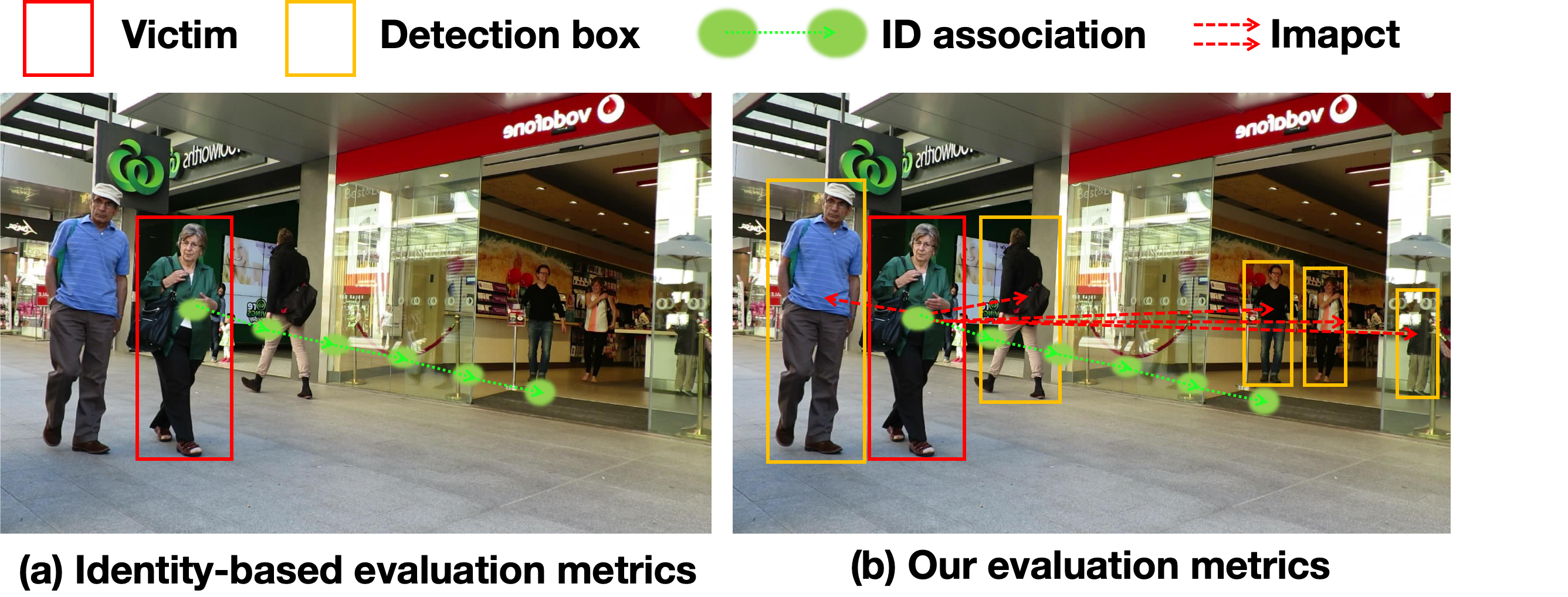}
\caption{A comparison of existing and our proposed evaluation metrics for MOT attacks. (a) Previous evaluation metrics are identity-based and only consider ID association. 
(b) Our proposed metrics emphasize the attack evaluation of both detection and ID association.
}
\label{Fig:evaluation metrics comparison} 
\end{figure}

As depicted in Figure \ref{Fig:evaluation metrics comparison}(a), identity-based evaluation metrics~\cite{hijack, lin2021trasw} evaluate the identity of the victim in every frame, which considers tracking. However, existing MOT algorithms typically include both tracking and detection modules. Therefore, while evaluating attacks on MOT, it is essential to calculate the performance of both tracking and detection modules simultaneously. 
Furthermore, identity-based evaluation metrics usually adopt a strategy that aims to attack one or a few specific targets. This approach overlooks the cumulative impact of attacking these specific targets on the entire system. For instance, as shown in Figure \ref{Fig:evaluation metrics comparison}(b), an attack on the victim (in the red detection box) might also affect the detection of other pedestrians (in yellow boxes), leading to switches of their identities or a decrease in detection accuracy. 

To overcome these limitations, we construct a maximum-attack-threshold evaluation metric that integrates detection and tracking. This type of  evaluation metric is based on the classic evaluation metrics for MOT~\cite{bernardin2008evaluating}. It provides a comprehensive evaluation of the robustness of MOT systems.

\paragraph{Tracking Attack Success Rate (TASR).} 
To assess the impact of specific targets during attacks on the overall system, TASR determines the change in MOTA and IDF1 before and after attacks and then divides it by the percentage of bounding boxes that are attacked ($R_{bbox}$). A high TASR indicates that the MOT system is highly susceptible to particular attacks, enabling an attacker to degrade its performance by compromising fewer targets. A low TASR implies that the attacker must compromise a greater number of targets to inflict considerable damage. The formula for TASR is:
$TASR = (MOTA + IDF1)_{decline}/2R_{bbox}$,
where $R_{bbox}$ is the ratio of the number of bounding boxes under attack to the total bounding boxes.  

\paragraph{Identity Obfuscation Rate (IOR).} 
To measure the effectiveness of adversarial attacks to cause identity obfuscation during data association in MOT, we define the Identity Obfuscation Rate as: 
$IOR =  D_{s} / P_{t}, \ \mathrm{where} \  P_{t} = N/T$,
where $T$ represents the number of frames that are attacked consecutively, $N$ is the total number of frames of the video,  $P_{t}$ denotes the maximum number of identity switches in the video, and $D_{s}$ is the number of identity switches throughout the video sequences.

\paragraph{Specific Target Attack Success Rate (STASR).} 
To evaluate attacks on specific targets in terms of both false positive rate (FP) and the number of identity switches, STASR focuses not only on how the attack interferes with the detection of the targets but also on how the attack disrupts the tracking of the targets. STASR is defined as:
$STASR = (FP + ID_{switch})_{increase} / A_{max}$,
where $A_{max} = P_{t} + P_{n}$, $P_{t}$ represents the maximum number of identity switches in the video, and $P_{n}$ denotes the number of detection boxes that are attacked. 

\section{Experiments}
In this section, we perform extensive empirical analysis of the proposed PapMOT attack on MOT in both digital and physical environments. We also conduct ablation studies on the adversarial loss and variants of adversarial patches. 

\subsection{Experimental Setup}

\paragraph{Datasets and Metrics.} 
We evaluate our attack on four popular MOT benchmarks: MOT15~\cite{MOT15}, MOT17~\cite{milan2016mot16}, MOT20~\cite{MOT20}, and BDD100K~\cite{bdd100k}. The target objects of the first three datasets are pedestrian-only, while BDD100K further includes vehicle objects.
We adopt our proposed evaluation metrics, namely TASR, SATSR, and IOR, to assess the performance of our PapMOT against state-of-the-art methods on the MOT task. More comparisons are found in Appendix.

\paragraph{Baselines.} 
Existing MOT attack methods can be divided into three categories: 1) methods that attack detectors in the MOT system by injecting pixel-level noise, \eg, RAnoise~\cite{guassian} and TOG~\cite{TOG}.
2) methods that attack detectors in the MOT system by placing optimized patches, e.g.,AdvTexture~\cite{advtexture} and ObjPatch ~\cite{thys2019fooling}. 
3) methods that attack both detectors and trackers in the MOT system by injecting pixel-level noise, e.g., Hijack~\cite{hijack} and Trasw~\cite{lin2021trasw}. 
Our PapMOT introduces a physically realizable adversarial patch into the MOT system to attack both detectors and trackers.

%

\paragraph{Victim Models.} We choose two representative MOT methods as victim models in our experiments: ByteTrack~\cite{bytetrack} that represents the tracking-by-detection paradigm and FairMOT~\cite{fairmot} that represents the joint detection and tracking paradigm. Besides, ByteTrack is equipped with YOLOX~\cite{yolox}, a more advanced version of the YOLO detector family trained on 
CrowdHuman~\cite{shao2018crowdhuman}, MOT17~\cite{milan2016mot16}, CityPerson~\cite{Shanshan2017CVPR}, and ETHZ~\cite{eth_biwi_00534} datasets. The pre-trained model of FairMOT is provided by PaddleDetection~\cite{ppdet2019}, and is trained on  Caltech Pedestrian~\cite{dollarCVPR09peds}, CityPersons~\cite{Shanshan2017CVPR}, CUHK-SYSU~\cite{xiaoli2017joint}, PRW~\cite{zheng2017person}, ETHZ~\cite{eth_biwi_00534}, and MOT17~\cite{milan2016mot16} datasets. 

\paragraph{Implementation Details.} 
In our digital experiments, to strike a balance between the stealthiness and effectiveness of attacks, we set the noise standard deviation of RAnoise to 10, and limit the maximum disturbance range in TOG to 8/255. The patch sizes for both AdvTexture and ObjPatch are set to one-third of the targeted detection box's size. We define an object as ``attacked'' if it overlaps with another object with an IOU greater than 0.2. 
In our physical experiments, we print the patches on foam boards with a fixed width and height of 50cm. The size ratio between the patch and an adult who is 1.6m tall is kept at around 30\%, ensuring that essential areas such as the face are not obstructed to maintain physical feasibility. For hyperparameters in the overall adversarial loss (\ie, Eq.~\ref{total loss}), we set $\beta$=1, $\gamma$=2.5, and $\delta$=2.
Datasets used in our patch generation consist of ETHZ, MOT17, and MOT17 clip datasets.

\begin{table*}[t]
\caption{Comparison with other representative adversarial attack methods on the MOT15 and MOT17 datasets. }
\resizebox{0.98\textwidth}{!}{
\small
\begin{tabular}{@{}ccccccccccccc@{}}
\toprule
 Dataset & \multicolumn{6}{c}{MOT15} & \multicolumn{6}{c}{MOT17} \\ 
\midrule
Victim & \multicolumn{3}{c}{FairMOT} & \multicolumn{3}{c}{ByteTrack}  & \multicolumn{3}{c}{FairMOT} & \multicolumn{3}{c}{ByteTrack} \\
\cmidrule(lr){1-1} \cmidrule(lr){2-4} \cmidrule(lr){5-7} \cmidrule(lr){8-10} \cmidrule(lr){11-13} 
Metrics & TASR & SATSR & IOR & TASR & SATSR & IOR & TASR & SATSR & IOR & TASR & SATSR & IOR \\
\midrule
RAnoise~\cite{guassian} & \ \ 0.7 & \ 0.6 & - & \ 0.6 & \ 0.3 & - & \ 0.3 & \ 0.5 & - & - & \ 0.2 & - \\
TOG~\cite{TOG} & 13.7 & \ 7.0 & \ 4.1 & 11.4 & \ 6.3 & \ 3.8 & \ 9.1 & \ 4.8 & \ 2.5 & \ 4.7 & \ 3.9 & \ 1.2 \\ 
\midrule
\ AdvTexture~\cite{advtexture} & 36.4 & 35.6 & \ 4.6 & 38.1 & 43.7 & \ 5.8 & 45.4 & 32.1 & \ 5.1 & 42.2 & 47.7 & \ 6.0 \\
\ ObjPatch~\cite{thys2019fooling} & 45.2 & 40.6 & \ 6.3 & 43.7 & 59.3 & \ 8.2 & 61.3 & 49.5 & \ 6.2 & 61.4 & 76.7 & \ 7.4 \\ 
\midrule
\ Hijack~\cite{hijack} & 14.2 & 18.5 & 39.3 & 11.4 & 15.9 & 29.0 & 10.0 & 19.2 & 26.5 & \ 8.4 & 15.4 & 25.4 \\
\ Trasw\cite{lin2021trasw} & 40.6 & 45.8 & \textbf{66.2} & 31.4 & 39.2 & 55.4 & 22.1 & 28.9 & 44.8 & 21.0 & 26.8 & 42.3 \\
\midrule
\ Ours & \textbf{76.4} & \textbf{85.0} & 62.3 & \textbf{85.3} & \textbf{79.2} & \textbf{65.9} & \textbf{76.8} & \textbf{93.0} & \textbf{46.2} & \textbf{94.6} & \textbf{89.6} & \textbf{79.5} \\ \bottomrule
\end{tabular}
}
\label{tab:experiments on MOT15 and MOT17}
\end{table*}

\subsection{Digital Domain Performance}
Table~\ref{tab:experiments on MOT15 and MOT17} and Table~\ref{tab:experiments on MOT20 and BDD100K} compare the  performance of various attacks on the MOT15, MOT17, MOT20, and BDD100K datasets. 
As can be seen, PapMOT achieves the highest attack success rate across all four datasets with a 50\%  bounding box attack, manifesting its superiority over existing methods (\ie, RAnoise~\cite{guassian}, TOG~\cite{TOG}, ADVTexure~\cite{advtexture}, ObjPatch~\cite{thys2019fooling}, Hijack~\cite{hijack}, Trasw~\cite{lin2021trasw}) against MOT in both pedestrian (\ie, MOT15, MOT17, and MOT20) and autonomous driving scenarios (\ie, BDD100K). We notice that the IOR of PapMOT is marginally below that of Trasw on the MOT15 and MOT20 datasets, which is possibly due to Trasw's sole emphasis on attacking data association. Nonetheless, PapMOT surpasses Trasw by a huge margin in terms of TASR and SATSR.
Overall, Trasw as a digital method is hard to be implemented in the physical world, and its attack is strictly triggered only when an object overlaps with the attack object. Thus, PapMOT outperforms Trasw in terms of functionality and overall attack performance. 

The superior performance of PapMOT is most evident on the MOT17 dataset, where it consistently surpasses all competing methods across all three metrics when applied against both FairMOT and ByteTrack. The success of PapMOT is largely due to incorporating the MOT17 clip dataset and MOT17 into its patch training dataset (more details about the clip dataset are found in Appendix).
%
It is noteworthy that TASR and SATSR both incorporate the evaluation of detection box performance, whereas IOR predominantly assesses identity association across frames. As such, it is observed that TASR and SATSR are typically higher for patch-based attacks and digital attacks compared to IOR (\eg, when attacking ByteTrack in MOT17, the IOR of AdvTexture is 6.0, while TASR and SATSR in ADvTexture are 42.2 and 47.7, respectively). This observation can be accounted for by the fact that these attacks primarily target detectors, which significantly affect the performance of detection boxes, but minimally impact identity association across frames.

\subsection{Physical Domain Performance}
While our patches have shown outstanding performance in the digital domain, translating this success to the physical world is challenging due to the ever-changing of physical environments.
In this section, 
we qualitatively evaluate the effectiveness of our adversarial patches, 
and further probe their limits under diverse physical environments.

\begin{table*}[t]
\caption{Comparison with other representative adversarial attack methods on the MOT20 and BDD100K datasets. }
\resizebox{0.98\textwidth}{!}{
\small
\begin{tabular}{@{}ccccccccccccc@{}}
\toprule
 Dataset & \multicolumn{6}{c}{MOT20} & \multicolumn{6}{c}{BDD100K} \\ 
\midrule
Victim & \multicolumn{3}{c}{FairMOT} & \multicolumn{3}{c}{ByteTrack}  & \multicolumn{3}{c}{FairMOT} & \multicolumn{3}{c}{ByteTrack} \\
\cmidrule(lr){1-1} \cmidrule(lr){2-4} \cmidrule(lr){5-7} \cmidrule(lr){8-10} \cmidrule(lr){11-13} 
Metrics & TASR & SATSR & IOR & TASR & SATSR & IOR & TASR & SATSR & IOR & TASR & SATSR & IOR \\
\midrule
RAnoise~\cite{guassian} & \ 2.6 & \ 0.9 & \ 0.4 & \ 4.1 & \ 1.5 & \ 0.4  & \ 2.2 & \ 0.7 & - & \ 1.9 & \ 0.7 & - \\
TOG~\cite{TOG} &  \ 8.6 & \ 2.5 & \ 0.6 &  \ 7.6 & \ 2.3 & \ 0.6 &  18.1 & \ 6.7 & \ 2.4 & 13.5 & \ 4.8 & \ 2.2 \\ \midrule
AdvTexture~\cite{advtexture}  & 23.5 & 18.5 & \ 2.9 &  32.0 & 26.7 & \ 3.2 &  21.1 & 12.6 & \ 4.4  & 18.3 & 19.0 & \ 4.8 \\
ObjPatch~\cite{thys2019fooling} &  42.3 & 21.4 & \ 4.4 & 41.3 & 27.5 & \ 6.8 & 26.3 & 19.4 & \ 5.2 & 20.5 & 23.3 & \ 7.6 \\ \midrule
Hijack~\cite{hijack} &  17.9 & 18.4 & 36.8 &  15.9 & 34.7 & 27.0 &  \ 7.1 & \ 9.8 & 10.5 &  \ 7.4 & \ 8.2 & 13.5 \\
Trasw~\cite{lin2021trasw} &  31.2 & 33.5 & \textbf{67.0}  & 26.6 & 28.5 & \textbf{64.2} &  12.2 & 17.4 & 23.3  & 11.0 & 13.1 & 19.3 \\ \midrule
Ours & \textbf{61.9} & \textbf{49.5} & 56.6  & \textbf{58.5} & \textbf{41.6} & 53.9  & \textbf{46.4} & \textbf{54.7} & \textbf{34.1} & \textbf{63.4} & \textbf{61.2} & \textbf{38.8} \\ \bottomrule
\end{tabular}
}
\label{tab:experiments on MOT20 and BDD100K}
\end{table*}

\begin{figure}[t]
\centering
\includegraphics[scale=0.29]{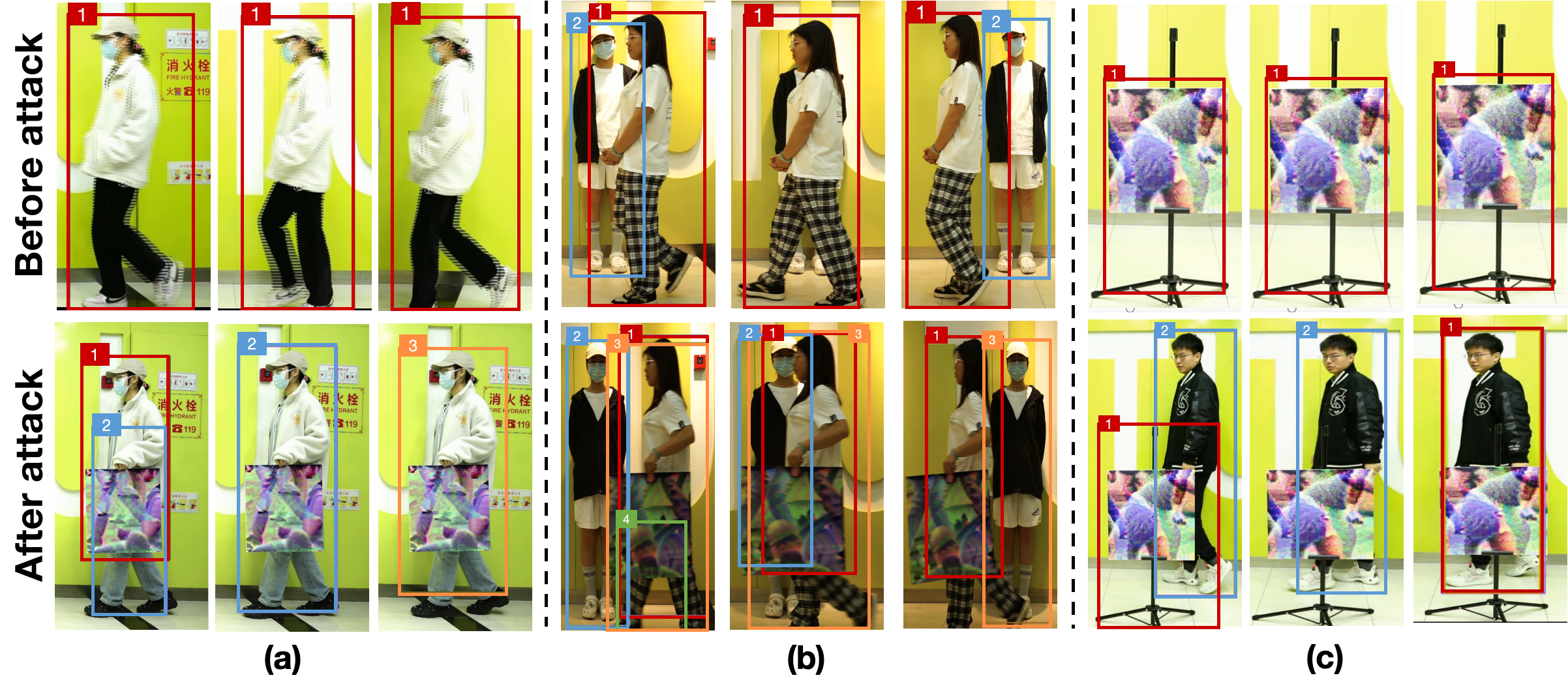}
\caption{Black-box transfer attack of different PapMOT patches in the real world. It illustrates the effective attack for the scenarios of a single pedestrian (a), two pedestrians overlap (b), and a pedestrian walking past from behind a fixed patch (c).}
\label{demonstrations of PapMOT in read worlds}
\end{figure}

\paragraph{Physical Validation.}
We train three patches with different datasets (\ie, ETHZ, MOT17, MOT20) for FairMOT and evaluate them against the ByteTrack detector. 
Figure~\ref{demonstrations of PapMOT in read worlds} showcases the demonstration of these adversarial patches for physical-world attacks under diverse scenes (refer to Supplementary Materials for a video demo).  

In Figure~\ref{demonstrations of PapMOT in read worlds} (a), an adversarial patch is attached to a pedestrian, targeting a single object. In the absence of our attack, the target's ID detected by the tracker is consistently 1, and the detected bounding box is accurate. When the attack is imposed, the detector produces an additional false detection box (in blue), and the accuracy of the initially accurate detection box deteriorates significantly. Moreover, the tracker is quickly deprived of the ability to identity associations across frames, and an ID switch takes place within 30 frames. These outcomes validate the effectiveness of our attack in single-object scenarios.

Figure~\ref{demonstrations of PapMOT in read worlds} (b) illustrates a multi-object scenario, where the adversarial patch is attached to one of the two pedestrians in the scene. When the pedestrian in white with the patch (denoted as ``target A'') passes by in front of the other subject in black (``target B''), their detected identities are successfully maintained in the absence of the attack. When attacked, however, the tracker mistakenly predicts two identities (\ie, 1 and 3) for target A. Also, an ID switch occurs and the accuracy of both boxes drops as target A passes by. This example shows our proposed attack works in multiple-object scenarios and its effect applies beyond the object with the attached patch. 

Figure~\ref{demonstrations of PapMOT in read worlds} (c) shows the scenario where a patch is kept stationary and is misidentified as a pedestrian by the tracker.  When attacked, a pedestrian passing behind it immediately causes an ID switch to occur (\ie, ID changes from 2 to 1). This example suggests that our adversarial patch is also suitable for static placement and is capable of affecting pedestrians as they pass by.

\begin{figure}[t]
\centering
\includegraphics[scale=0.34]{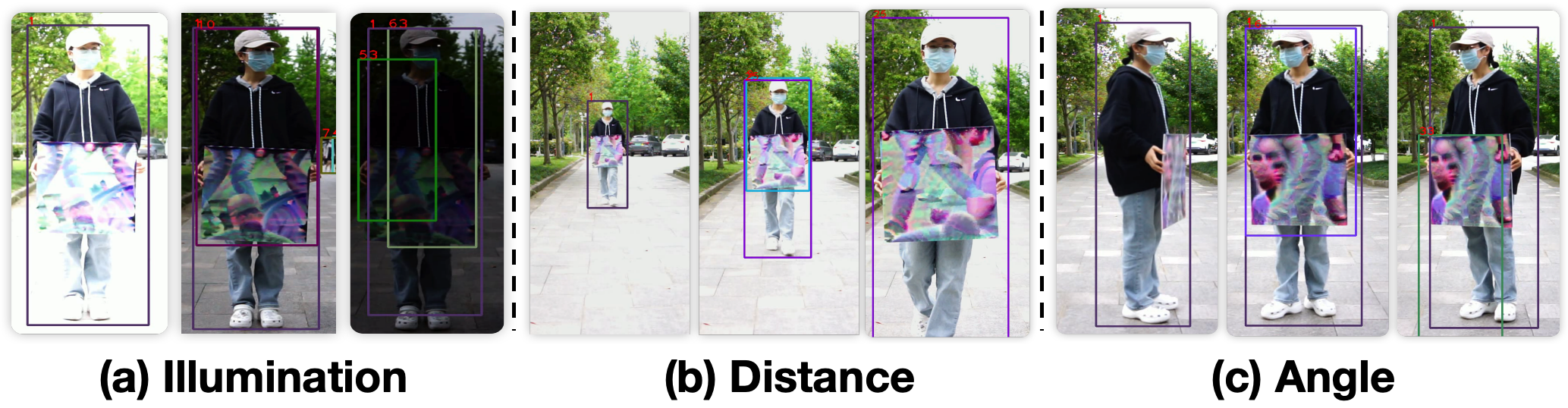}
\caption{Studies on patch effectiveness at varying illuminations, distances, and angles.}
\label{Patch effectiveness testing under different physical conditions.}
\end{figure}

\paragraph{Attack Boundary.}
We examine three key physical conditions that may affect the effectiveness of patch attacks: (a) illumination, (b) distance, and (c) angle, shown in Fig.~\ref{Patch effectiveness testing under different physical conditions.}. 
For illumination, we sweep over the entire spectrum from total darkness to total brightness by adjusting the ISO sensitivity of the camera. We divide the illumination levels into 10 segments between 0 and 1. An illumination value of 0.5 denotes the normal exposure under overcast outdoor conditions with the ISO set to 200; 0 and 1 correspond to full darkness and full brightness, respectively.
For distance, we vary the range between the patch and the camera from 0.5 to 10 meters to obtain a better view.
For angle, we control the angle of the patch relative to the camera by adjusting the deflection angle of the person and using 5-degree intervals to present the final attack results. The quantitative evaluation results of the attack boundary effectiveness for three different patches are presented in Table~\ref{bounds of effectiveness of PapMOT}. 

Based on the experimental results, several important conclusions can be drawn. Firstly, different patches can achieve varying attack effects depending on the physical conditions. Secondly, the distance between the patches and the camera must be maintained within a specific range (0.8m-4.1m) to achieve an effective attack. Thirdly, patch attacks work better in relatively dark environments, resulting in more false detection boxes. Moreover, a larger patch size has a stronger attack ability. However, to ensure the stealthiness and effectiveness of the patch, the patch size in PapMOT is set to approximately 30\% of the target size. Lastly, to ensure effective attack outcomes, the angle of the patch relative to the camera generally needs to be limited to a small range (\ie, within 15 degrees). It may be considered to add rotation transformation in the patch training to enhance the robustness of the patches against rotation.

\begin{table}[t]
\caption{Studies on the effectiveness range of three adversarial patches.}
\centering
\begin{tabular}{ccccc}
\toprule
&   Distance   &   Illumination   &   Patch Size   &  Angle  \\
\midrule
\ Patch 1 & 0.8m-3.1m & 0.4-0.7 & 40cm-100cm & \quad 0°-25° \quad \\
\ Patch 2 & 1.2m-4.1m & 0.5-0.8 & 40cm-100cm & \quad 0°-15° \quad \\
\ Patch 3 & 1.1m-3.6m & 0.4-0.8 & 40cm-100cm & \quad 0°-30° \quad \\ 
\bottomrule
\end{tabular}
\label{bounds of effectiveness of PapMOT}
\end{table}

\subsection{Ablation Studies}
We analyze the effectiveness of each adversarial loss term in our proposed training objective. We also conduct real-world experiments to evaluate the robustness against unoptimized patches, including real images and uniform color patches.


\paragraph{Loss Functions.}
%
Table \ref{tab:Effectiveness analysis of the losses on MOT17 dataset} shows how different adversarial losses affect the performance of our proposed on against MOT17. The results indicate that PapMOT without $\mathcal{L}_{bbr}$ performs significantly worse than PapMOT without $\mathcal{L}_{ap}$ in terms of IOR, scoring 7.2 and 52.6, respectively, when attacking ByteTrack. This suggests that the $\mathcal{L}_{bbr}$ loss is critical in disrupting the consistency of data association. Conversely, when attacking FairMOT, PapMOT without $\mathcal{L}_{ap}$ scores substantially lower in TASR and SATSR than PapMOT without $\mathcal{L}_{bbr}$, with scores of 11.0 and 13.1, respectively, compared to 50.5 and 43.3. This underscores the significant impact of the $\mathcal{L}_{ap}$ loss on detector performance.

\begin{figure}[t]
\centering
\includegraphics[scale=0.33]{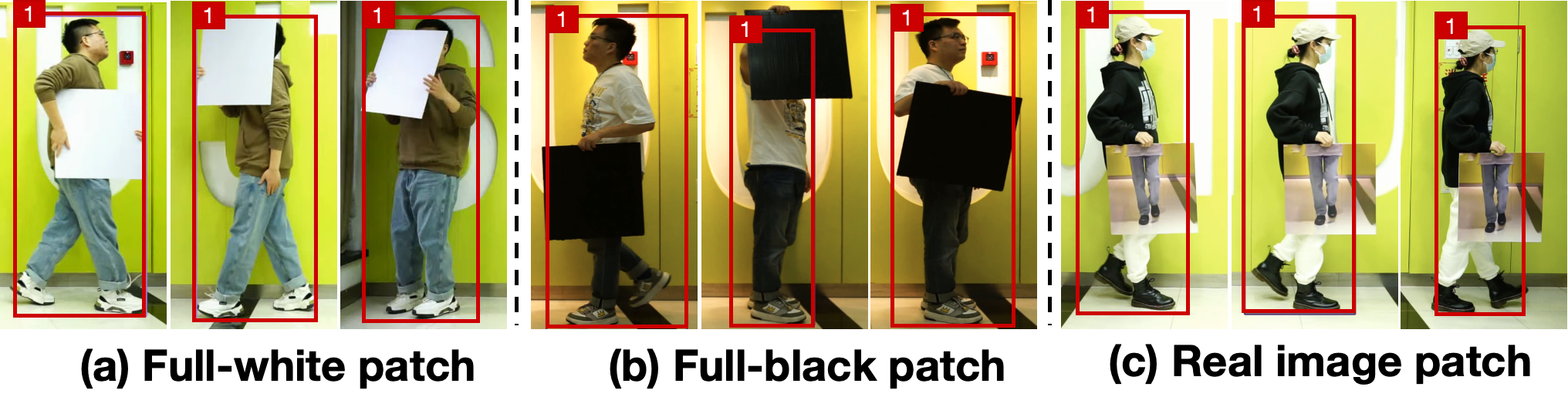}
\caption{The attack testing using real image, full-white, and full-black patches on the MOT system. These patches have the same size of 50$\times$50cm as ours.}
\label{white and real pic}
\end{figure}

\begin{table}[t]
\caption{The effectiveness analysis of the adversarial losses.}
\centering
\begin{tabular}{ccccc}
\toprule
\multirow{2}{*}{Victim } & \multirow{2}{*}{\quad \quad Attacker} & \multicolumn{3}{c}{Metrics} \\ 
\cmidrule(l){3-5} 
 &  & TASR & SATSR & IOR \\
\midrule
\multirow{3}{*}{\ \ ByteTrack~\cite{bytetrack}\ \ } & PapMOT w/o $\mathcal{L}_{bbr}$ & 61.3 & 64.7 & 7.2 \\
 & \ \ PapMOT w/o $\mathcal{L}_{ap}$ \ \ & 21.9 & 26.8 & 52.6 \\
 & PapMOT & \textbf{94.6} & \textbf{89.6} & \textbf{79.5} \\ 
\midrule
\multirow{3}{*}{FairMOT~\cite{fairmot}} & PapMOT w/o $\mathcal{L}_{bbr}$ & 50.5 & 43.3 & 5.6 \\
 & PapMOT w/o $\mathcal{L}_{ap}$ & 11.0 & 13.1 & 29.3 \\
 & PapMOT & \textbf{76.8} & \textbf{93.0} & \textbf{46.2} \\ 
\bottomrule
\end{tabular}
\label{tab:Effectiveness analysis of the losses on MOT17 dataset}
\end{table}


\paragraph{Various Patches.}
We evaluate the effectiveness of attacking the MOT system using unoptimized patches (\eg, full-white patches, full-black patches, and real image patches). As shown in Figure \ref{white and real pic}, although the target is partially occluded by (a) the full-white patch or (b) the full-black patch, it is tracked steadily and the ID remains unchanged throughout the whole video sequence. Similarly, although the target is partially occluded by  (c) the real-image patch, it is tracked steadily and the ID is maintained at 1 throughout the entire video sequence. The experiments indicate that it is difficult to attack MOT systems with adversarial patches that are not optimized.

\section{Conclusion}
In this paper, we introduce a novel physical adversarial attack method named PapMOT against multiple object tracking systems. Unlike all previous attacks that inject noise into input video frames in the digital domain, our approach optimizes printable adversarial patches that effectively disrupt the data association of MOT trackers across frames in the actual physical world. We demonstrate the feasibility and effectiveness of PapMOT in both digital and physical scenarios through extensive experiments on multiple datasets. In future work, we plan to investigate the creation of more natural-looking adversarial patches, such as transforming these patches into clothing to deceive the MOT system. Additionally, we will expand physical adversarial MOT attacks from pedestrian scenarios to autonomous driving scenarios.

\noindent \textbf{Acknowledgements.} This work was supported in part by NSFC (62322113, 62376156).



%
%
\bibliographystyle{splncs04}
\bibliography{main}
\end{document}